# Transformers Improve Breast Cancer Diagnosis from Unregistered Multi-View Mammograms

Xuxin Chen [1,*], Ke Zhang [1], Neman Abdoli [1], Patrik W. Gilley [1], Ximin Wang [2], Hong Liu [1], Bin Zheng [1], and Yuchen Qiu [1,*]

1   School of Electrical and Computer Engineering, University of Oklahoma, Norman, OK, USA 73019;
2   XPeng Motors, Guangzhou, Guangdong, China 510640;
*   Correspondences: xuxin.chen@ou.edu, qiuyuchen@ou.edu

**Abstract:** Deep convolutional neural networks (CNNs) have been widely used in various medical imaging tasks. However, due to the intrinsic locality of convolution operation, CNNs generally cannot model long-range dependencies well, which are important for accurately identifying or mapping corresponding breast lesion features computed from unregistered multiple mammograms. This motivates us to leverage the architecture of Multi-view Vision Transformers to capture long-range relationships of multiple mammograms from the same patient in one examination. For this purpose, we employ local Transformer blocks to separately learn patch relationships within four mammograms acquired from two-view (CC/MLO) of two-side (right/left) breasts. The outputs from different views and sides are concatenated and fed into global Transformer blocks, to jointly learn patch relationships between four images representing two different views of the left and right breasts. To evaluate the proposed model, we retrospectively assembled a dataset involving 949 sets of mammograms, which include 470 malignant cases and 479 normal or benign cases. We trained and evaluated the model using a five-fold cross-validation method. Without any arduous preprocessing steps (e.g., optimal window cropping, chest wall or pectoral muscle removal, two-view image registration, etc.), our four-image (two-view-two-side) Transformer-based model achieves case classification performance with an area under ROC curve (AUC = 0.818±0.039), which significantly outperforms AUC = 0.784±0.016 achieved by the state-of-the-art multi-view CNNs (p = 0.009). It also outperforms two one-view-two-side models that achieve AUC of 0.724±0.013 (CC view) and 0.769±0.036 (MLO view), respectively. The study demonstrates the potential of using Transformers to develop high-performing computer-aided diagnosis schemes that combine four mammograms.

**Keywords:** Transformer; mammogram; multi-view; self-attention; computer-aided diagnosis; breast cancer; classification; deep learning

## 1. Introduction

Breast cancer is one of the most common cancers diagnosed in women [1]. Mammography is recommended as the standard imaging tool to perform population-based breast cancer screening. Although early detection using screening mammography can help reduce breast cancer mortality, mammogram interpretation is a difficult task for radiologists, as reflected by the low diagnostic yield (e.g., detecting 3.6 cancers are per 1000 (0.36%) mammography screenings) [2] and the relatively high false positive recall rates (around 10%) [3]. To help radiologists detect and diagnose breast cancer more accurately and reduce large inter-reader variability, computer-aided detection/diagnosis (CAD) schemes have been developed aiming to provide radiologists "a second opinion" [4-6].

With the help of machine learning technologies, different types of CAD schemes of mammography have been developed over the last two decades [7-9]. Since 2015, deep learning (DL) has become the mainstream technology in developing mammographic CAD schemes due to its state-of-the-art performance in the laboratory studies [10-12]. DL-based approaches have the advantage of processing a whole image to learn useful feature representations, without the need of manually extracting features. However, most of existing mammographic CAD schemes are single view image-based schemes, while in a typical mammography screening, each patient has four mammograms acquired from two views namely, craniocaudal (CC) and mediolateral oblique (MLO) view of the left (L) and right (R) breasts. Radiologists read these 4 images simultaneously. Thus, researchers have well recognized that developing multi-view or multi-image-based CAD schemes had significant advantages over single image-based CAD schemes [13, 14]. Recently, applying deep learning models to develop multi-view CAD schemes of mammograms has also attracted research interest. For example, Carneiro and colleagues performed multi-view mammogram analysis by finetuning ImageNet-pretrained convolutional neural networks (CNNs) to classify between malignant and benign microcalcification cluster and soft tissue mass like lesions [15, 16]. Although mammograms of the CC and MLO view were not registered, their multi-view models outperformed single-view models (based on CC or MLO) by a large margin. In another study, Wu and colleagues further explored how to combine the features from different views and breast sides more effectively [17]. They proposed a "view-wise" feature merging approach that employed two *ResNet-22* models. One *ResNet-22* takes two CC images as input and concatenates the LCC and RCC feature representations for malignancy prediction, while the other model is used to process LMLO and RMLO images. The proposed approach made separate predictions for CC and MLO view images, and the predictions were averaged during inference.

The methods introduced above are based on CNNs. In the most recent literature, Vision Transformers [18] are quickly emerging as a strong alternative to CNNs-based architectures in a variety of medical imaging tasks, such as image classification [19, 20] and segmentation [21, 22]. One core concept of Transformers is the so-called self-attention mechanism [23]. It can help models to adaptively learn "what" and "where" to attend to, so that a subset of pertinent image regions or features will be used to facilitate performing the target task [24]. Enabled by the self-attention mechanism, Vision Transformers have shown advantages at capturing long-range dependencies of the input sequence.

For multi-view mammograms, two types of dependencies are important: within-mammogram dependency and inter-mammogram dependency. The within-mammogram dependency refers to pixel/patch relationships in a mammogram. For example, an image patch that contains a suspicious mass is more likely to receive more attention than other patches, due to differences in mammographic density, shape, texture, etc. Meanwhile, this patch may be more closely related to its surrounding patches than the image patches at a far distance. These different kinds of patch relationships within a mammogram can be learned by Transformers to identify suspicious patches and predict likelihood of the case being malignant more accurately. As for the inter-mammogram dependency, it focuses on the pixel/patch relationships between mammograms of different sides or views. Accordingly, inter-mammogram dependencies can be further divided into two categories. On the one hand, bilateral feature difference between the left and right mammograms (e.g., LCC vs. RCC) of the same view can be learned, which is important because in clinical practice, it is common to see that either one breast shows malignancy, or both breasts are benign/negative, whereas the situation that both the left and right breasts are malignant rarely occurs. Especially for a malignant case, the left and right mammograms tend to exhibit different characteristics in terms of mammographic density, mass presence, etc. Such bilateral feature difference has been shown useful in delivering more accurate results for near-term breast cancer risk prediction [25-27] and malignancy identification [28]. On the other hand, Transformers can highlight the correspondence between the ipsilateral mammogram patches (e.g., LCC and LMLO) coming from the same breast. For instance, if a suspicious mass is present in LCC image, it should also be seen in LMLO image, and vice versa.

Despite Transformers' great potential in modeling long-range dependencies, their applications in multi-view mammogram analysis remain largely unexplored. At the time of writing, only one study so far has partially leveraged Transformers for this purpose [29]. The authors applied and tested a hybrid model (Transformer and CNNs) by using one of two global cross-view Transformer blocks to fuse intermediate feature maps produced by CC and MLO views, so that feature representations from one view can be transferred to the other. In comparison, our study has two unique characteristics. First, our model is basically a pure Transformer instead of a hybrid Transformer and CNN model. This provides us the convenience to easily and efficiently employ pretrained Transformers with as few architectural modifications as possible. Second, we exploit both local and global transformer blocks within one Transformer model to learn within-mammogram dependencies and inter-mammogram-dependencies, respectively. Thus, to the best of our knowledge, this is the first CAD study in which a whole Transformer with both local and global blocks is integrated for multi-view mammogram analysis, through implicitly learning different mammographic dependencies.

## 2. Materials and Methods

### 2.1 Materials

From an existing de-identified retrospective full-field digital mammography (FFDM) image database pre-assembled in our medical imaging research laboratory, we assembled a dataset that contains 3,796 mammograms acquired from 949 patients for this study. Specifically, each patient has four mammograms acquired from CC and MLO view of the left and right breasts, which are named as LCC, RCC, LMLO, and RMLO view images, respectively. All mammograms were acquired using Hologic Selenia digital mammography machines (Hologic Inc, Bedford, Massachusetts, USA), which have a fixed pixel size of 70μm. Depending on breast sizes, the acquired mammograms have two sizes of either 2558×3327 or 3327×4091 pixels.

Among all the patients, 349 were diagnosed as screening negative or benign (BIRAD 1 or 2) by radiologists, and the rest were recalled and recommended to do biopsy due to the detection of suspicious soft-tissue masses. Based on the histopathological reports of tissue biopsy, 130 and 470 cases were confirmed to be benign and malignant, respectively. To perform binary classification using Transformers, we grouped all screening negative cases and benign lesion cases in one class of benign (or cancer-free), while the cases confirmed having malignant lesions are grouped into another class of cancer or malignant cases. Since in this study, we only focused on predicting the likelihood of the mammographic case being malignant, the original FFDM images were subsampled using a pixel averaging method using a kernel size of 5×5 pixels. As a result, the sizes of two types of original FFDM images are reduced to 512×666 and 666×819 pixels, respectively, with the pixel size of 0.35mm. More detailed information of our FFDM image database have been reported in our previous studies (i.e., [30]).

### 2.2 Methods

Figure 1 shows the proposed pipeline of using Vision Transformers for breast cancer detection from the unregistered four multi-view mammograms. Most similar to our work is the Multi-view Vision Transformer (MVT) [31] for 3D objects recognition in natural images . In this study, we adopt a similar design as MVT to detect breast cancer depicting on mammograms. MVT has two major parts namely, the local Transformer blocks and global Transformer blocks. The local Transformer blocks process information from each view image (e.g., LCC, RCC, LMLO, RMLO) individually. In comparison, the global Transformer blocks process information from the four view mammograms jointly. The local and global Transformer blocks share the same design, and their key components include self-attention, multi-head attention and multi-layer perceptron. We will first introduce these components in the following subsections. After that, we will describe our model's inputs (i.e., patch and position embeddings) and provide more details about local and Transformer blocks.

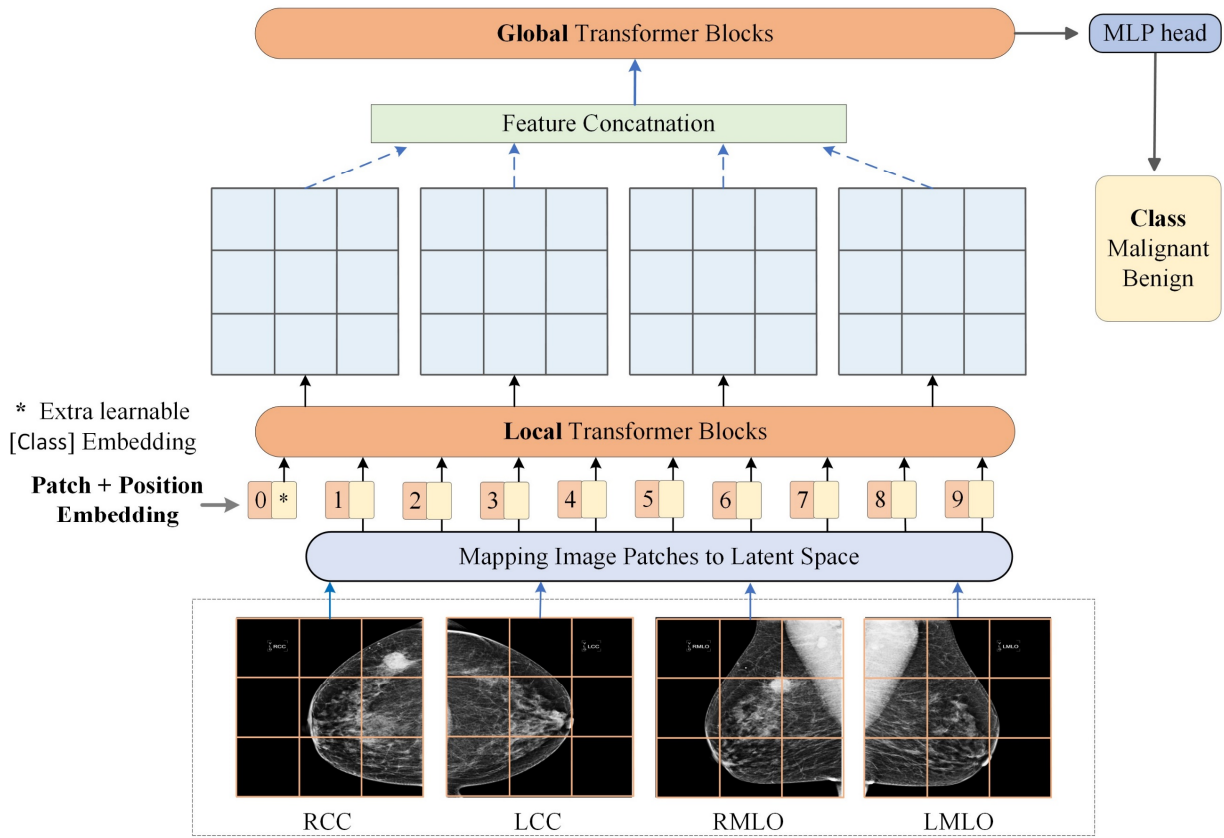

**Figure 1.** Overview of using Transformers for multi-view mammogram analysis. For a patient, each of LCC, RCC, LMLO, and RMLO mammogram is split into image patches and mapped to embedding vectors in the latent space. Then, positional embeddings are added to patch embeddings. The sequence of embedding vectors is sent into local Transformer blocks to learn within-mammogram dependencies. Weights of local Transformer blocks are shared among the four mammograms. The four outputs are concatenated into one sequence and fed into global Transformer blocks to learn inter-mammogram dependencies. The class token of the last global Transformer block is sent into an MLP head to classify the case as benign/malignant.

### 2.2.1 Transformers background

**Self-attention (SA)**

When processing a patch at a certain position, the SA mechanism simultaneously determines how much focus to place on patches at other positions. For each image patch embedding $z$ of dimension $d_{embed}$, the inputs of an SA module are three vectors, namely query and key of dimension $d_k$, and value of dimension $d_v$. They can be obtained by multiplying the patch embedding by their respective weight matrices $W_q, W_k, W_v$, all of which are trainable parameters as computed using Equation (1).

$$q = zW_q, \quad k = zW_k, \quad v = zW_v \qquad (1)$$

To speed up the computation for a sequence of image patch embeddings **z**, the queries, keys, and values vectors are packed into the Query ($Q$), Key ($K$), and Value ($V$) matrices. In Vision Transformers, SA can be computed using Equation (2) and is often implemented in the form of "Scaled Dot-Product Attention" [23], as shown in Figure 2(b).

$$\text{SA}(Q, K, V) = \text{Softmax}\left(QK^T / \sqrt{d_k}\right)V \qquad (2)$$

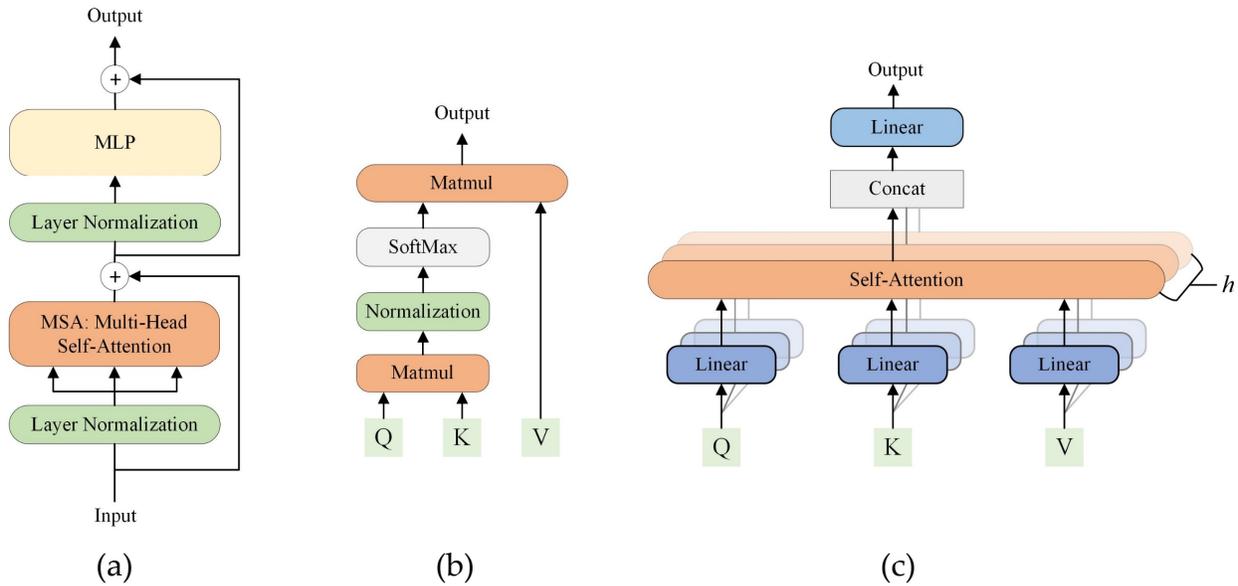

Figure 2. (a) The Transformer block consists of an MSA, an MLP, skip connections, and layer normalizations. (b) Self-attention (Scaled Dot-Product Attention). Matmul: multiplication of two matrices. (c) Multi-head attention consists of multiple parallel self-attention heads. Concat: concatenation of feature representations. $h$: the number of self-attention heads.

Given a query (one row of $Q$), the dot products of this query and all keys ($K^T$) are computed, resulting in a score vector that measures the similarities between that query and different keys. Scaling the similarity scores by a factor $\sqrt{d_k}$ helps to stabilize the training process. Then a softmax layer normalizes these similarity scores into positive weights that range from 0 to 1. The weights control the amount of attention that different values should receive. Finally, multiplication of the weight vector and $V$ produces an output for the query.

**Multi-head self-attention (MSA)**

At the core of Transformers is the MSA mechanism [18, 23]. Unlike SA that executes attention operations once, MSA adopts multiple SA layers (also known as *heads*) to perform *h* self-attention operations in parallel. As shown in Figure 2(c), the input queries, keys, and values are first linearly projected into different subspaces for *h* times. Then, the projections of $Q$, $K$, and $V$ are sent into their corresponding SA head. Finally, the outputs from all heads are concatenated and projected again, yielding an output for the input queries. The formulas to compute *heads* and MSA can be expressed as Equations (3) and (4).

$$head_i = \text{SA}\left(QW_i^Q, KW_i^K, VW_i^V\right) \quad (3)$$

$$\text{MSA}(Q, K, V) = \text{Concat}(head_1, \ldots, head_h)W^O \quad (4)$$

where $W_i^Q$, $W_i^K$, $W_i^V$, $W^O$ are trainable parameter matrices used for linear projections. In practice, each head has $d_k = d_v = d_{embed}/h$. MSA is beneficial to Transformers design because it drives the model to jointly learn and ensemble information from different representation subspaces. This is somewhat reminiscent of using large amounts of stacked kernels in CNNs to capture as many useful features as possible.

**Multi-layer perceptron (MLP)**

In each Transformer block (Figure 2(a)), following MSA is an MLP module that consists of two fully-connected layers with a GELU non-linearity [32]. The first layer of MLP transforms its input to a higher-dimensional space, while the second layer restores the dimension back to be same as the input. Besides, both the MSA and MLP modules have a residual connection [33], followed by layer normalization [34].

### 2.2.2 Transformers for malignancy identification from multi-view mammograms

**Patch and position embedding**

Patch embedding [18] aims at mapping raw image patches to a new representation space, in which it is easier for models to process information. Given a 2D mammogram image $x \in R^{H \times W \times C}$, it is divided into smaller image patches of resolution $(P, P)$ and then flattened, resulting in a sequence of image patches $x_P \in R^{N \times (P \times P \times C)}$, where $N = HW/P^2$ is the number of image patches, $(H, W)$ and $C$ denote the image resolution and number of channels. This sequence is then mapped to a latent space of dimension $d_{embed}$. The mapping is normally implemented as a trainable linear projection or convolutions, denoted by $\boldsymbol{E}$. In this study, $\boldsymbol{E}$ is a convolutional layer with kernel size equal to patch size, while the original MVT model employs a fully-connected layer. The mapping's outputs are referred to as *patch embeddings*. Meanwhile, a *classification token* $x_{class}$, which is an extra learnable vector, is prepended to the sequence of patch embeddings. Therefore, the sequence has $(N + 1)$ elements, and each element is a $d_{embed}$-dimensional vector.

Since patch embeddings do not carry much absolute or relative position information that may be important for models to achieve satisfactory performance, *positional embeddings* are performed to preserve such spatial information. Learnable 1D position embeddings ($E_{pos}$) are implemented and added to patch embeddings of the input sequence. Thus, the input sequence embeddings of LCC, RCC, LMLO, RMLO mammogram can be expressed using Equation (5) as below.

$$\mathbf{z}_{LCC} = [x_{class}; x^1_{P\_LCC}\mathbf{E}; x^2_{P\_LCC}\mathbf{E}; \dots ; x^N_{P\_LCC}\mathbf{E}] + \mathbf{E}_{pos}$$

$$\mathbf{z}_{RCC} = [x_{class}; x^1_{P\_RCC}\mathbf{E}; x^2_{P\_RCC}\mathbf{E}; \dots ; x^N_{P\_RCC}\mathbf{E}] + \mathbf{E}_{pos}$$

(5)

$$\mathbf{z}_{LMLO} = [x_{class}; x^1_{P\_LMLO}\mathbf{E}; x^2_{P\_LMLO}\mathbf{E}; \dots ; x^N_{P\_LMLO}\mathbf{E}] + \mathbf{E}_{pos}$$

$$\mathbf{z}_{RMLO} = [x_{class}; x^1_{P\_RMLO}\mathbf{E}; x^2_{P\_RMLO}\mathbf{E}; \dots ; x^N_{P\_RMLO}\mathbf{E}] + \mathbf{E}_{pos}$$

It should be noted that the class token $x_{class}$, patch mapping $\mathbf{E}$, and positional embeddings $\mathbf{E}_{pos}$ are shared for different mammogram views. This differs from the original MVT that separately devises a class token for each view.

**Local Transformer blocks**

By denoting the total number of local Transformer blocks as $L_{local}$, the sequence embeddings from each view image goes through the local Transformer blocks sequentially as presented in Equation (6).

$$\mathbf{z}^0 = \{\mathbf{z}^0_{LCC}, \mathbf{z}^0_{RCC}, \mathbf{z}^0_{LMLO}, \mathbf{z}^0_{RMLO}\} = \{\mathbf{z}_{LCC}, \mathbf{z}_{RCC}, \mathbf{z}_{LMLO}, \mathbf{z}_{RMLO}\},$$

$$\mathbf{z}^l_{\sim} = \text{MSA}\left(\text{LN}(\mathbf{z}^{l-1})\right) + \mathbf{z}^{l-1}, \qquad l = 1 \dots L_{local} \qquad (6)$$

$$\mathbf{z}^l = \text{MLP}\left(\text{LN}(\mathbf{z}^l_{\sim})\right) + \mathbf{z}^l_{\sim}, \qquad l = 1 \dots L_{local}$$

where LN represents layer normalization. Note that we use "$\{\mathbf{z}_{LCC}, \mathbf{z}_{RCC}, \mathbf{z}_{LMLO}, \mathbf{z}_{RMLO}\}$" to denote sequential operation. In other words, the respective sequence embeddings of four mammograms from the same patient are processed by local Transformer blocks sequentially. As shown in Figure 2(a), the first local Transformer block takes a sequence $\mathbf{z}^0$ of dimension $(1, N + 1, d_{embed})$ as input. For the first block ($l = 1$), the embedded sequence goes through layer normalization and MSA (output $\mathbf{z}^1_{\sim}$), another layer normalization and MLP (output $z^1$). The output of the former Transformer block ($l-1$) is continuously used as the input of the next block ($l$) until $l = L_{local}$. As a result, through this block-by-block sequential operation, the final block outputs a sequence of the same dimension as shown in Equation (7).

$$z^{L_{local}} = \{z_{LCC}^{L_{local}}, z_{RCC}^{L_{local}}, z_{LMLO}^{L_{local}}, z_{RMLO}^{L_{local}}\} \qquad (7)$$

**Global Transformer blocks**

Then, by defining total number of global Transformer blocks as $L_{global}$, our model concatenates the last local block's outputs for the four mammograms ($z^{L_{local}}$) and automatically sends them into the global Transformer blocks. The concatenated sequence $g^0$ is of dimension $(1, 4N + 4, d_{embed})$. In this way, the global Transformer blocks can jointly attend to the four mammograms from the same patient. Due to Transformers' advantage in modeling long-range dependencies, we expect that the bilateral feature difference as well as ipsilateral mammogram correspondence can be effectively captured and leveraged toward improving the classification results. The computation sequence is expressed by Equation (8).

$$g^0 = \text{concat}(z_{LCC}^{L_{local}}, z_{RCC}^{L_{local}}, z_{LMLO}^{L_{local}}, z_{RMLO}^{L_{local}})$$

$$g_{\sim}^l = \text{MSA}\left(\text{LN}(g^{l-1})\right) + g^{l-1}, \qquad l = 1 \ldots L_{global} \qquad (8)$$

$$g^l = \text{MLP}\left(\text{LN}(g_{\sim}^l)\right) + g_{\sim}^l, \qquad l = 1 \ldots L_{global}$$

where $g_{\sim}^l$ and $g^l$ represent outputs of 1) layer normalization plus MSA processing and 2) layer normalization plus MLP processing in the sequence of global Transformer blocks ($l = 1 \ldots L_{global}$), respectively.

Last, the model used in this study is based on the DeiT architecture [35] that has 12 Transformers blocks stacked together. We vary the number of local and global Transformer blocks while keeping the total number the same (i.e., $L_{local} + L_{global} = 12$). Class token of the last global block is passed into a one-layer MLP (head) to perform mammogram classification. Our model follows the original DeiT design without introducing any new components, so that the pretrained weights of DeiT can readily be used to finetune the local, global blocks, and the MLP head.

### 2.2.3 Experiments

**Implementation details**

The original mammograms are 12-bit images with 4096 grayscale levels. To use the pretrained Transformer for finetuning, we duplicated and stacked the mammograms across the RGB channels and normalized the pixel values to be within [0, 255]. Meanwhile, all the mammograms were resized to be 224×224 pixels, and image patches are of size 16×16 pixels.

We implemented the model in PyTorch. The pretrained tiny (5.5 M parameters) and small (21.7 M parameters) DeiTs without distillation [35] were finetuned on our mammogram dataset. The original MLP head for classification was modified to have 2 outputs (malignant/benign). We kept all the default settings (e.g., learning rate, optimizer, etc.) for finetuning, except setting batch size as 8 and disabling the augmentations like MixUp. For five-fold cross-validation, we trained the model for 500 epochs, and most models were able to converge within 200 epochs.

All the experiments were performed using a single NVIDIA GeForce RTX 2080 Ti GPU with 11GB VRAM. Five-fold cross-validation was used to train and evaluate models. For the Transformers based on tiny DeiT, training a single-view-two-side (2-image), two-view-two-side (4-image) model took 10.1 and 19.5 hours, respectively. For the two-view-two-side (4-image) model based on small DeiT, the training time was 22.4 hours.

**Performance evaluation**

In summary, the new model is applied to process four view mammograms of one study case together and generate one prediction score that represents the probability or likelihood of the case depicting malignant lesion. The model-generated prediction scores are compared with biopsy-confirmed ground-truth to determine model detection and classification accuracy. In this study, we take two steps of data analysis methods to evaluate and compare model performance. First, we perform a standard receiver operating characteristic (ROC) type data analysis using a maximum likelihood-based ROC curve fitting program (ROCKIT, http://metz-roc.uchicago.edu/MetzROC/software) to generate ROC curves. The corresponding AUC value along with the standard deviation (STD) is computed as an index to evaluate model performance to classify between malignant and benign cases. The significant differences ($p$-values) between AUC values are also computed for comparing classification performance (AUC values) of different models. Second, since model has two outcome nodes representing two classes of case being malignant or benign, a testing case is signed to malignant class if the probability score of malignancy is greater than the probability score of benign or vice versa. After case assignment, we build a confusion matrix to record true-positive (TM), false-positive (FP), true-negative (TN) and false-negative (FN) of classification results. Then, for each model, we compute and report the average of validation accuracies from five folds and their standard deviations using Equations (9) and (10).

$$\overline{ACC} = \frac{1}{n}\sum_{i=1}^{n}\frac{TM_i + TN_i}{All\ Cases_i} \qquad (9)$$

$$STD = \sqrt{\frac{\sum_{i=1}^{n}(ACC_i - \overline{ACC})^2}{n}} \qquad (10)$$

where $n = 5$ in five-fold cross-validation.

In addition, from the confusion matrix, we also compute and compare several different evaluation indices of model performance, including precision, recall, specificity, F1 score. We compare our two-view Transformer to two other state-of-the-art CNN-based multi-view models as published in the literature [16, 17]. We let these two CNN models adopt the same ResNets-18 architecture [33]. The first CNN (named as "CNN feature concatenation") concatenates the feature representations obtained from four ResNets and sends them into a fully-connected layer to perform classification. The second CNN employs two ResNets for different views, and thus we name this model as "View-wise CNN". More details of this model can be found in the previous introduction section. Last, we also compare the two-view Transformer with two single-view Transformers.

## 3. Results

Table 1 shows and compares the trend of lesion patch detection and final case classification accuracy of the new MVT model when using different numbers of local blocks and global blocks. The model is applied to process unregistered four view mammograms and tested using a five-fold cross-validation method. The result demonstrates that by using a combination involving 2 local Transformer blocks and 10 global Transformer blocks, the new MVT model yields the highest case-based malignancy classification accuracy (77.0%) as well as the relatively higher reproducibility or robustness indicated by the smaller standard deviation (STD = ±1.2) than those yielded by most of other block combination options as shown in Table 1. The comparison results also show that using more global blocks to capture more details of feature relationship between four bilateral and ipsilateral mammograms plays a more important role than local blocks to detect and extract image features in one case of four images, in order to achieve higher lesion detection or classification accuracy. The comparison results show that the best-performing MVT model employs 2 local and 10 global Transformer blocks to detect breast cancer using four mammograms.

**Table 1.** Summary of model-generated case classification accuracy (ACC) along with the standard deviation (STD) in five-fold cross-validation on the unregistered, four-view mammogram dataset (LCC, RCC, LMLO, RMLO), with different numbers of local and global Transformer blocks (using a small DeiT model).

| Local blocks | 0 | 2 | 4 | 8 | 12 |
|---|---|---|---|---|---|
| Global blocks | 12 | 10 | 8 | 4 | 0 |
| Fold 1 | 79.5 | 78.9 | 78.9 | 73.7 | 73.2 |
| Fold 2 | 74.7 | 77.4 | 77.4 | 73.7 | 74.7 |
| Fold 3 | 74.7 | 76.3 | 76.3 | 72.6 | 68.4 |
| Fold 4 | 74.7 | 75.8 | 75.3 | 74.2 | 71.1 |
| Fold 5 | 73.5 | 76.7 | 73.5 | 73.2 | 74.6 |
| Mean ACC (%) ± STD | 75.4 ± 2.3 | **77.0 ± 1.2** | 76.3 ± 2.0 | 73.5 ± 0.6 | 72.4 ± 2.7 |

Figure 3 demonstrates six ROC curves representing six different classification models using four Transformer-based models including two bilateral single-view (CC or MLO) two-side mammograms using tiny DeiT architecture and two multi-view (two-view-two-side) mammograms using either the tiny DeiT or small DeiT architecture, respectively, as well as two CNN-based multi-view image models. Comparison of these ROC curves shows that using CC view images alone to finetune Transformers (tiny DeiT architecture) yields the lowest mammographic case classification performance (AUC = 0.724±0.013). However, once CC and MLO images are used together for finetuning, the classification performance is substantially improved with AUC = 0.814±0.026. Additionally, using our Transformer-based models that combine 4 images together also delivers the significantly higher classification performance (i.e., AUCs) than using two conventional CNN-based models ($p < 0.05$).

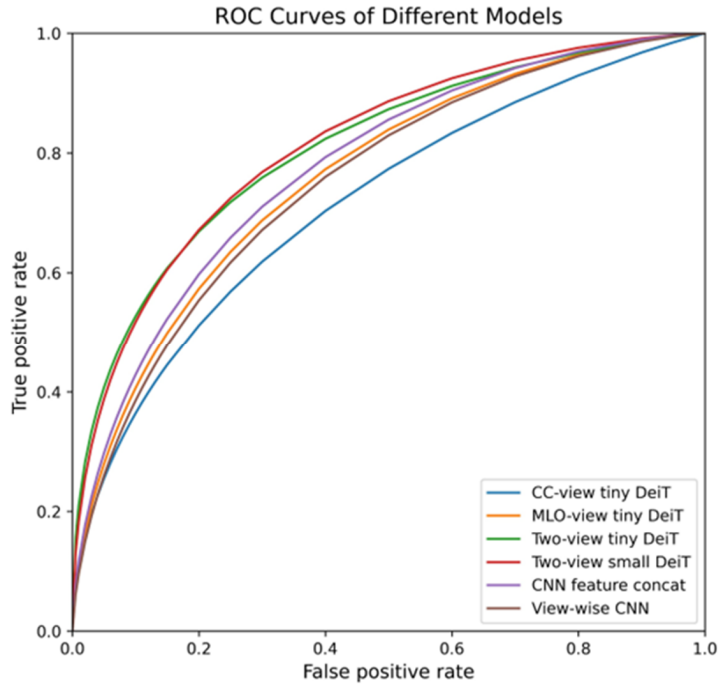

**Figure 3.** Illustration of 6 ROC curves of single-view Transformers, multi-view Transformers, and CNNs.

Figure 4 shows a confusion matrix generated by classification results of four image Transformer based model using small DeiT architecture. Table 2 summarizes 5 classification performance indices (accuracy, prevision, recall, specificity and F-1 score) computed from confusion matrix and AUC value computed from ROC curve. The table also compares classification performance of four models including two state-of-the-art models developed and published by other two groups of researchers using our image dataset and the same five-fold cross-validation method. The comparison results demonstrate several interesting results or observations.

First, the model (Single-view DeiT-tiny) has the smallest number of parameters that need to be finetuned using mammograms. As a result, it can not only improve model training or finetuning efficiency, but also help increase model robustness by reducing the risk of overfitting using relatively small medical image datasets.

Second, comparing with the use of only either CC or MLO view images, fusion of four mammograms of both CC and MLO view as input simultaneously to finetune the model yields significantly higher malignancy classification accuracy and/or AUC value ($p < 0.05$) with smaller standard deviations, which indicates that fusion of useful features computed from 4 images can help reduce image feature noise and irrelevance to the malignancy prediction task.

Third, the two-view models using tiny DeiT and small DeiT exhibit very close classification performance. This suggests that finetuning large-sized Transformers does not necessarily contributes to significant performance boost than finetuning small-sized

Transformers on small medical image datasets. Fourth, the performance of two-view Transformer-based models surpasses the two existing CNN-based models reported in previous studies [16, 17], when measured by a variety of performance evaluation indices. This is because Transformers can more effectively identify and fuse useful information from 4 images.

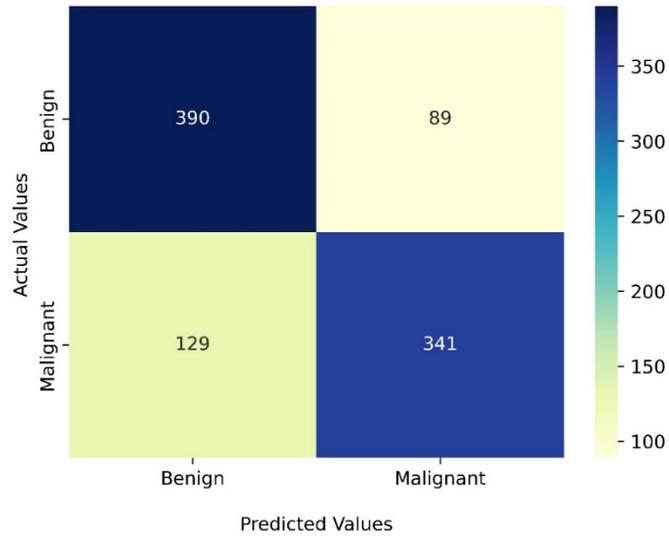

**Figure 4.** Confusion matrix of the two-view tiny DeiT's classification results.

**Table 2.** Result comparisons of the proposed model and other models. Params means the number of parameters. To compute $p$-values for different two-view models, the model proposed in paper [16] is used as the reference base.

| Model | Params | CC | MLO | Accuracy (%) | Precision | Recall | Specificity | F-1 score | AUC | $p$-value |
|---|---|---|---|---|---|---|---|---|---|---|
| Single-view DeiT-tiny | 5.5 M | ✓ | | 69.6±1.4 | 0.708±0.035 | 0.662±0.040 | 0.728±0.058 | 0.683±0.013 | 0.724±0.013 | <0.001 |
| Single-view DeiT-tiny | 5.5 M | | ✓ | 72.5±2.8 | 0.728±0.037 | 0.713±0.027 | 0.737±0.045 | 0.720±0.028 | 0.769±0.036 | 0.376 |
| Two-view DeiT-tiny (**proposed**) | 5.5 M | ✓ | ✓ | **77.0±1.2** | **0.797±0.039** | **0.726±0.063** | **0.814±0.057** | **0.757±0.022** | **0.814±0.026** | 0.031 |
| Two-view DeiT-small (**proposed**) | 21.7 M | ✓ | ✓ | 76.3±2.8 | **0.799±0.071** | 0.706±0.033 | **0.818±0.081** | 0.747±0.018 | **0.818±0.039** | 0.009 |
| CNN feature concatenation [16] | 44.7 M | ✓ | ✓ | 73.9±2.4 | 0.761±0.05 | 0.696±0.041 | 0.781±0.071 | 0.725±0.019 | 0.784±0.016 | - |
| View-wise CNN [17] | 22.4 M | ✓ | ✓ | 71.7±2.1 | 0.735±0.04 | 0.677±0.051 | 0.756±0.069 | 0.702±0.021 | 0.759±0.023 | N/A |

## 4. Discussion

Since mammograms are 2D projection images and breast tissue and lesions are quite heterogeneous in different bilateral and ipsilateral imaging views (CC and MLO view of two breasts), features extracted and computed from different view images are often quite different and may contain complementary lesion detection or classification information. However, developing multi-view image-based CAD schemes have not been successful to date due to many remaining technical challenges including difficulty of reliable image registration. In this paper, we present a novel study to develop multi-view image CAD scheme directly applying to un-registered four mammograms. The study has several unique characteristics and generates several new observations.

First, we recognize that within-mammogram dependencies and inter-mammogram dependencies are important for accurately predicting the likelihood of the mammographic case depicting malignant lesions. We also recognize that unlike human eyes that can easily identify and register lesions depicting on ipsilateral (CC and MLO) views and bilateral tissue asymmetry between left and right breast images, applying CAD schemes to register mammograms or lesions is very difficult due to the lack of reliable fiducials in mammograms. Thus, one of the significant advantages of this study is that we develop and test a new CAD scheme that does not require image registration. Four mammograms are fed into the MVT model "as is" directly. Due to the intrinsic capability in modeling long-range relationships of the input sequence, we exploit Transformers to capture different kinds of mammogram dependencies and/or associated features extracted from different images. We use local Transformer blocks to model patch relationships within each of the LCC, RCC, LMLO, RMLO mammograms individually and global Transformer blocks to capture patch relationships among these four mammograms jointly. Skipping traditional image registration step does not only increase efficiency of developing multi-view image-based CAD schemes, but also enhance robustness of CAD schemes due to avoid the impact of image registration errors.

Second, to verify the effectiveness of our proposed new method (Table 2), we compare our new two-view-two-side four-image based DeiT-tiny Transformer with two bilateral single-view models (either CC or MLO view). When using one bilateral sing-view mammograms of left and right breasts as input of the model, we observe that although both CC and MLO view mammograms contain useful discriminatory information for malignancy identification, bilateral MLO-view mammograms in our dataset seem to contain more valuable information, so that the model trained using MLO view images performs better than the model trained using CC view images (i.e., 0.769±0.036 vs. 0.724±0.013 in AUC values or 72.5±2.8% vs. 69.3±1.3% in accuracy as shown in Table 2), which is consistent with previous studies (i.e., [28]). However, when using four images of two-view-two-side (both CC and MLO views of two breasts) as input, the new two-view-two-side four image Transformer-based model can fuse the information extracted from four images and capture more meaningful features from one view that may be missed by the other to further improve case classification performance. Compared with two single-view models using either two bilateral CC or MLO view images, the two-view-two-side four-image Transformer-based model yields statistically higher classification accuracy with higher AUC (0.814±0.026) and accuracy (77.0 ± 1.2%). This aligns well with one of our early expectations or study hypothesis that the global Transformer blocks can effectively capture the correspondence of the image features extracted and computed from four different-view mammograms. Like the behavior of radiologists in reading mammograms by considering information extracted from four images together, optimal fusion of images from four mammograms also enables to help the new CAD model achieve higher performance or accuracy in classification between malignant and benign lesions or cases. Thus, this study provides new scientific data or evidence to further support the significance of developing multi-view image-based CAD schemes in the future.

Third, the state-of-the-art performance of CNNs for multi-view mammogram analysis has been strongly reliant on several key factors, such as cropping off background areas and using high-resolution images (e.g., 2677×1942 pixels) [17], obtaining breast mass or microcalcification masks [15, 16], etc. However, it is not that easy to satisfy these requirements and maintain scientific rigor or robustness of such CNN-based CAD schemes in practice. On the contrast, our new two-view Transformer-based model does not rely on any of these requirements. It does not need any image preprocessing steps such as optimal window cropping, chest wall or pectoral muscle removal in MLO view images. It works well on low-resolution images, with much smaller number of model parameters of (i.e., 5.5M in the Transformer model using DeiT-tiny architecture versus 44.7M in ResNets-18 feature concatenation). This is a significant advantage of applying our new model in the future clinical applications. We also recognize that performance of deep learning model-based CAD schemes reported in the literature vary significantly depending on the use of different image datasets and/or testing or cross-validation methods [36]. In order to demonstrate that this new CAD model can achieve higher or very comparable lesion detection and classification performance as current state-of-the-art CAD schemes, we compare performance of several models using the same image datasets and cross-validation method. As shown in Table 2, our new model yields the highest lesion classification accuracy. Thus, it is natural to speculate that if one or more of the complicated steps is performed, the two-view Transformer will have a performance boost. Especially, if breast mass annotations for the mammograms are incorporated into Transformer training, the single-view and two-view models should be able to learn important patch relationships more quickly, thus improving the malignancy classification performance.

Fourth, our multi-image Transformer-based model is basically a pure Transformer model, except that one convolutional layer is used for image patch embedding, and it also uses and combines multi-layer local and global Transformer blocks. Thus, our new model is very different from a recently reported study [29] that only replaced the global pooling methods used in CNN model by either a global image token-based or image pixel-based cross-view Transformer and reported the highest classification performance of AUC = 0.803±0.007 using token-based cross-view Transformer applying to an old digitized screen-film mammogram dataset (DDSM). Although due to the use of different image datasets, the classification performance of our new model (AUC = 0.818±0.039) is not directly comparable to the previous study [29], we believe that experiment data analysis results of this new study is promising to demonstrate the feasibility and advantages of using a new model that combines both local and global block Transformers to optimally or effectively extract, match and fuse clinically relevant image features from four mammograms to develop CAD schemes of multi-view mammograms. Despite the promising results, we also recognize that this is the first preliminary study using this pure Transformer model with both local and global block Transformers, more comprehensive studies are needed. In the future, it is also worthy exploring a hybrid architecture of CNNs and Transformers, since the feature maps produced by intermediate CNN layers may also carry some low-level details that may provide complementary information or features to help further improve lesion detection and/or malignancy classification.

## 5. Conclusions

Vision Transformers are quickly emerging as powerful architectures for learning long-range dependencies in the last two years. However, their potential in the field of multi-view mammogram analysis remains largely unleashed to date. In this study, we propose using local and global Transformer blocks to model within-mammogram and inter-mammogram dependencies, respectively. The model is successfully applied to process four view mammograms simultaneously. By comparing with other state-of-the-art multi-view image-based CAD schemes, our new model yields higher lesion classification performance with small variance (or standard deviation). Therefore, the promising results demonstrate that Multi-view Vision Transformers (MVTs) are a strong or better alternative to CNNs toward building high-performing and robust multi-view image-based CAD schemes of FFDM images. The reported study results should be further validated using new large and diverse image datasets in the future studies.

**Author Contributions**: Conceptualization, X.C. and X.W.; methodology, X.C.; software, X.C.; validation, X.C., K.Z. and Y.Q.; formal analysis, K.Z.; investigation, X.C.; resources, B.Z.; data curation, X.C.; writing—original draft preparation, X.C.; writing—review and editing, B.Z, and Y.Q; visualization, N.A. and P.G.; supervision, H.L and Y.Q.; project administration, H.L.; funding acquisition, H.L and B.Z. All authors have read and agreed to the published version of the manuscript.
**Funding:** This research was funded in part by National Institutes of Health, USA, under grant number P20 GM135009.

Institutional Review Board Statement: This study does not acquire new image data from the hospitals. It uses the existing de-identified image data selected from a pre-established retrospective database in our medical imaging laboratory.

**Data Availability** Statement**:** For the detailed information of the image data, please contact the corresponding author, Dr. Xuxin Chen.

**Conflicts of Interest:** The authors declare no conflict of interest.